\documentclass[sigconf]{acmart}

\acmConference[SIGIR '21] {Proceedings of the 44th International ACM SIGIR Conference on Research and Development in Information Retrieval}{July 11--15, 2021}{Virtual Event, Canada.}
\acmBooktitle{Proceedings of the 44th International ACM SIGIR Conference on Research and Development in Information Retrieval (SIGIR '21), July 11--15, 2021, Virtual Event, Canada}
\acmPrice{15.00}
\acmISBN{978-1-4503-8037-9/21/07}

\copyrightyear{2021}
\acmYear{2021}
\setcopyright{acmcopyright}
\acmDOI{10.1145/3404835.3462981}

\acmSubmissionID{1143}

\usepackage{graphicx}
\usepackage{float}
\usepackage{subfigure}
\usepackage{amsmath}
\usepackage[misc]{ifsym}
\usepackage{lipsum}
\usepackage{balance}

\begin{document}
\fancyhead{}
\title{LPF: A Language-Prior Feedback Objective Function for De-biased Visual Question Answering}

\author{Zujie Liang, Haifeng Hu\textsuperscript{$\ast$}, Jiaying Zhu}
\affiliation{%
  \institution{School of Electronics and Information Technology, Sun Yat-sen University}
  \country{}
}
\email{{liangzj9,zhujy53}@mail2.sysu.edu.cn, huhaif@mail.sysu.edu.cn}








\renewcommand{\shortauthors}{Zujie Liang et al.}

\begin{abstract}
Most existing Visual Question Answering (VQA) systems tend to overly rely on the language bias and hence fail to reason from the visual clue. 
To address this issue,
we propose a novel Language-Prior Feedback (LPF) objective function, to 
re-balance 
the proportion of each answer's loss value in the total VQA loss.
The LPF firstly calculates a modulating factor to determine the language bias using a question-only branch. Then, the LPF assigns a self-adaptive weight to each training sample in the training process. With this reweighting mechanism, the LPF ensures that the total VQA loss can be reshaped to a more balanced form. By this means, the samples that require certain visual information to predict will be efficiently used during training. Our method is simple to implement, model-agnostic, and end-to-end trainable. We conduct extensive experiments and the results show that the LPF (1) brings a significant improvement over various VQA models, (2) achieves competitive performance on the bias-sensitive VQA-CP v2 benchmark.
\end{abstract}


\vspace{-1em}


\maketitle

\newcommand\blfootnote[1]{%
\begingroup
\renewcommand\thefootnote{}\footnote{#1}%
\addtocounter{footnote}{-1}%
\endgroup
}

\begin{figure}[!t]
  \centering
  \includegraphics[width=\linewidth]{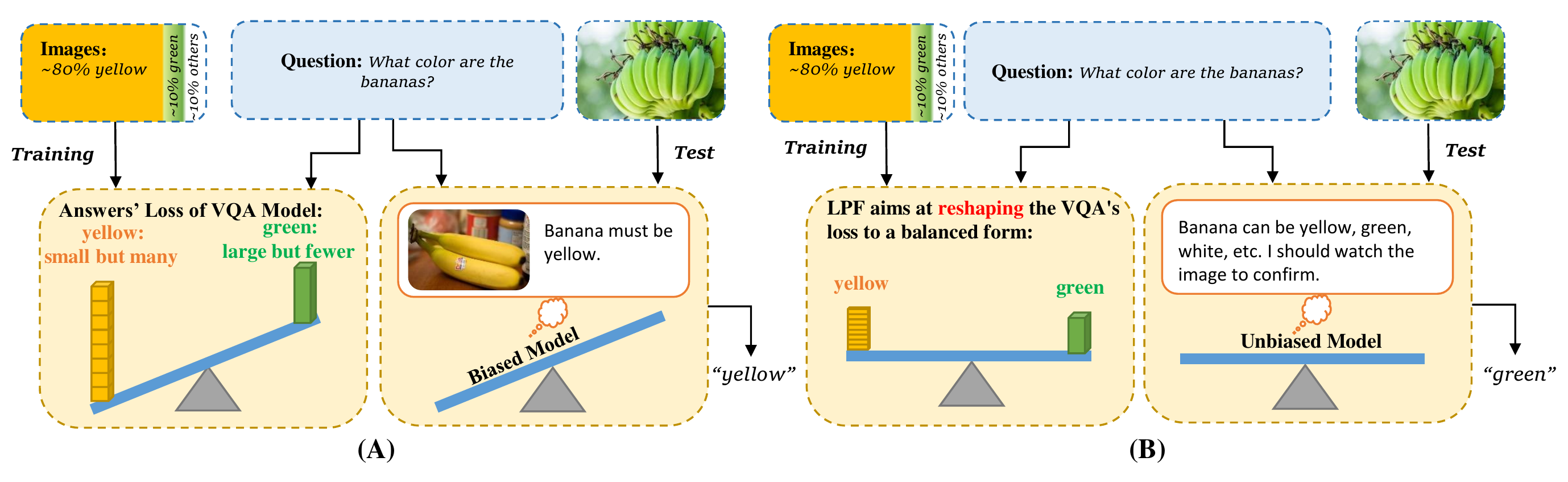}
  \caption{An example of how the original extremely imbalanced answer distribution and the LPF impact the learning process of VQA models. 
  \vspace{-1em}
  }
  \label{fig:figure1}
\end{figure}
\section{Introduction}
Visual Question Answering (VQA), which aims to answer questions about the given visual content, is one of the most challenging multi-modal tasks involving natural language processing and computer vision~\cite{Antol_2015_ICCV}. Nowadays, the state-of-the-art VQA models~\cite{Anderson_2018_CVPR,jiang2018pythia,Tan2019LXMERTLC,Shah_2019_CVPR,wu2019generating,Cadene_2019_CVPR,NIPS2018_7429} can achieve well performance on the famous benchmarks~\cite{Antol_2015_ICCV,Goyal_2017_CVPR}. However, recent studies~\cite{Selvaraju_2019_ICCV,guo2019quantifying,Agrawal_2018_CVPR} demonstrate that most existing VQA models overly leverage superficial correlations between question and answer words to generate answers, \emph{i.e.} language priors, resulting in poor performance in real-world settings. For instance, about 80\% of questions “what color are the bananas?" have the answer “yellow" in the training set, the VQA models tend to “assume" that the answer should be “yellow" without “looking" at the given image. To overcome these language prior problems in VQA, ~\cite{Agrawal_2018_CVPR} introduced VQA-CP dataset, which has different QA distributions in the training and test splits. This dataset is proposed to evaluate the generalization performance of VQA models. Most state-of-the-art VQA models that learned language biases from training data perform poorly in the VQA-CP.

\blfootnote{\textsuperscript{$\ast$}Corresponding author. This work was supported in part by the National Natural Science Foundation of China under Grant 62076262.}
Currently, the widely used solutions to mitigate the bias can be roughly grouped into annotation-based methods and non-annotation based methods. 
The annotation-based methods try to align the VQA models' visual attention to the human attention, yet require expensive human annotation~\cite{Selvaraju_2019_ICCV,NIPS2019_9066}. Not only this, recent work~\cite{shrestha-etal-2020-negative,DBLP:conf/nips/TeneyAKSKH20} shows that the accuracy improvements stem from a regularization effect rather than proper visual grounding. 
Besides, counterfactual data augmentation techniques~\cite{zhu2020overcoming,chen2020counterfactual,abbasnejad2020counterfactual,teney2020learning,liang2020learning} are proposed to balance the training data and boost the generalization performance of the model.
However, how to make the VQA models generalize well under imbalanced training data still remains a major challenge.
Another prevailing way of non-annotation based methods is training a question-only model to regularize the training of the VQA model, which can further be grouped into two types: 1) \textit{adversary-based}~\cite{NIPS2018_7427,grand2019adversarial}: they train the question-only branch to prevent the question encoder from capturing unwanted language priors in an adversarial training scheme. But these methods bring significant noise and make the training instable~\cite{grand2019adversarial}. 2) \textit{ensemble-based}~\cite{NIPS2019_8371,clark2019don,karimi-mahabadi-etal-2020-end,niu2020counterfactual}: they use the ensemble strategy to combine the two models' prediction, which derives the training gradients based on the fused answer distributions. Still and all, current ensemble-based methods merely increase or decrease the loss value of each sample independently, 
the global training objective can still be unbalanced across different answers. 
Since the ground truth answer distribution per question type is extremely long-tailed~\cite{Agrawal_2018_CVPR}, as shown in Fig.~\ref{fig:figure1}, tons of well-classified samples (with small loss values) can still construct a total loss with non-trivial magnitude, and subsequently dominate the gradient of the training process, which is not the original intention of training. We posit that it is of crucial importance to re-balance the proportion of each answer's loss value in the total VQA loss, rather than simply increase or decrease its loss value. 

To address this issue, we propose a reweighting-based method, namely Language-Prior Feedback (LPF) objective function. 
Concretely, the LPF calculates a modulating factor to determine the language bias using the output distribution of a question-only branch. Then, a dynamic weight is assigned to each training sample in the training process. 
It is worth noting that, the reweighting mechanism ensures that the total VQA loss can be reshaped to a more balanced form (as one example shown in Fig.~\ref{fig:banana_example}), which is different from the ensemble-based methods such as RUBi~\cite{NIPS2019_8371}. 
By optimizing the LPF objective function and question-only model simultaneously, the VQA model is encouraged to focus on the samples that require certain visual knowledge to predict. 
Not only that, the LPF method is easy to implement, model-agnostic and end-to-end trainable, which is compatible with most current state-of-the-art VQA systems. 

Overall, the contributions of this paper are summarized as follows: 
(1) We introduce a Language-Prior Feedback objective function (LPF) that automatically reshapes the training loss to a balanced form when learning from the imbalanced VQA dataset, 
and propose a generic framework leveraging a question-only branch to calculate a dynamic weight for different answers.\footnote{Codes: https://github.com/jokieleung/LPF-VQA}
(2) Extensive experiments are conducted on the popular
language bias sensitive benchmark VQA-CP v2. The results validate that LPF brings significantly performance gain over strong baseline and achieves competitive performance against the state-of-the-art methods.

\section{Methodology}

\begin{figure}[t]
  \centering
  \includegraphics[width=\linewidth]{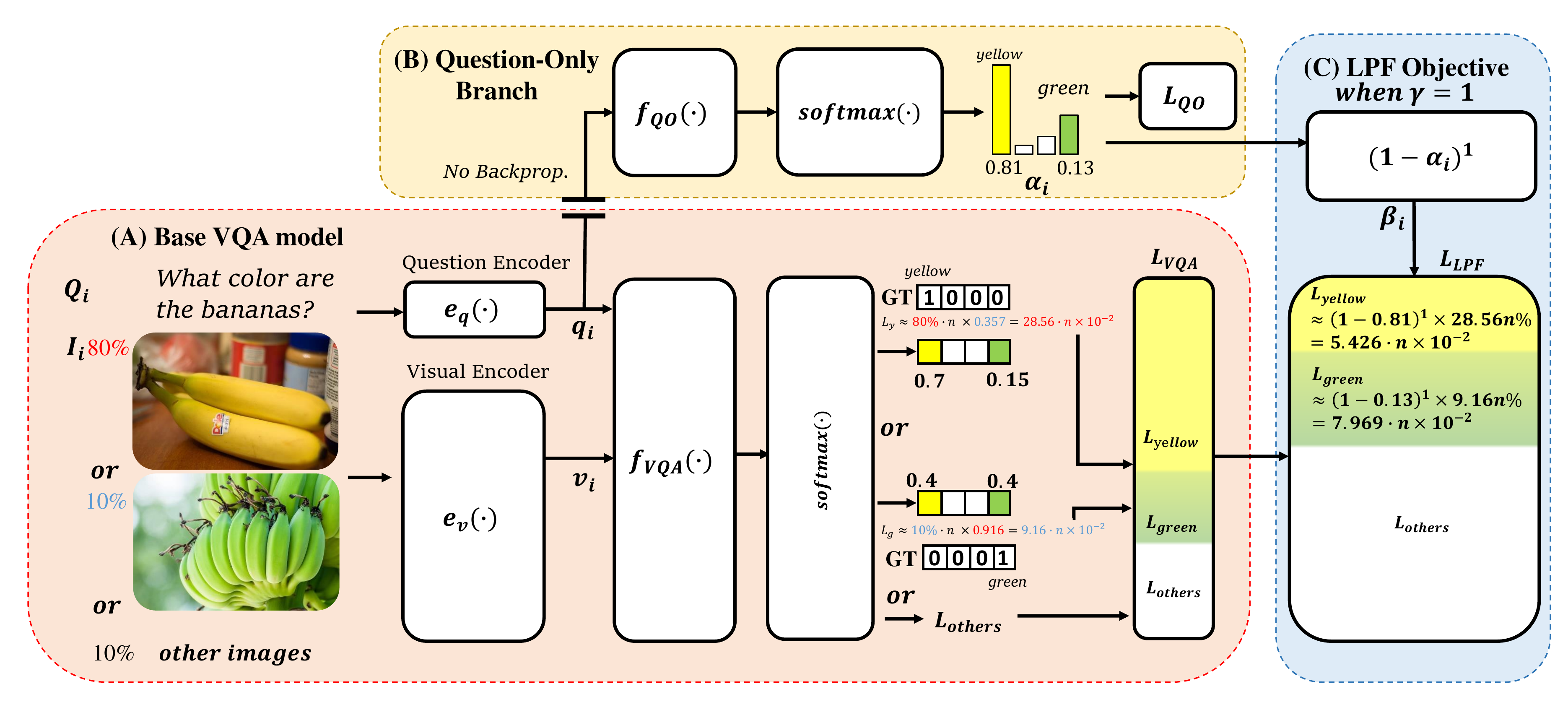}
  \caption{The overview of our framework and detailed illustration of the LPF impact on the training process. $n$ in this example is the number of images corresponding to the given question. The classical biased VQA training objective is reshaped to a balanced form by the LPF.
  \vspace{-1em}
  }
  \label{fig:banana_example}
\end{figure}

In this section, we introduce our approach named Language-Prior Feedback (LPF) objective. In summary, the LPF estimates the intensity of the model bias to language modality through a question-only model and dynamically tunes the loss weight of each training sample so as to train an unbiased model. The overall structure of our model together with a specific example is shown in Fig.~\ref{fig:banana_example}, which includes three modules: (1) An arbitrary base VQA model (2) A question-only branch (3) A reshaped VQA objective function.

\subsection{An arbitrary VQA model}

Most of previous VQA systems~\cite{Anderson_2018_CVPR,NIPS2018_7429,benyounes2019block} consider the Visual Question Answering task as a multi-class classification problem. In general, they produce feature representations for the given image and question, then fuse them into a multi-modal representation with bilinear fusion methods~\cite{NIPS2018_7429,Ben-younes_2017_ICCV}, attention mechanisms~\cite{Anderson_2018_CVPR,Gao_2019_CVPR}, etc. 
Finally, the multi-modal representation is ultilized to predict a distribution over the answer space.
Formally, given a set consisting of $ N $ triplets of images $I_{i} \in \mathcal{I}$, question $Q_{i} \in \mathcal{Q}$ and answer $a_{i} \in \mathcal{A}$, we denote as $\mathcal{D}=\left\{I_{i}, Q_{i}, a_{i}\right\}_{i=1}^{N}$. 
For each input image $\mathit{I_i}$, a visual object embedding vector set $\mathbf{v_i}=\left\{v_1, v_2, ..., v_k\right\}$ are extracted by an object detector $e_v$, \emph{i.e.},$\mathbf{v_i} = e_v(I_i)$. 
For each input question $\mathit{Q_i}$, it is encoded sequentially using a question encoder (such as LSTM, GRU, etc.) denoted by $e_q$ to produce a question embedding vector, \emph{i.e.},$\mathbf{q_i} = e_q(Q_i)$. 
Further, the final answer distribution is calculated by a mapping $f_{VQA}\left(\mathbf{v}_{\mathbf{i}}, \mathbf{q}_{i}\right)$.
The classical VQA training objective of minimizing the standard cross entropy loss can be formalized as follows:
 \begin{equation}
  \label{con:loss_vqa}
\begin{aligned}
    \mathcal{L}_{VQA}
    =-\frac{1}{N} \sum_{i=1}^{N} \log \left(softmax\left(f_{VQA}\left(\mathbf{v}_{\mathbf{i}}, \mathbf{q}_{i}\right)\right)\right) \left[a_{i}\right]
\end{aligned}
\end{equation}

\subsection{Determining Language Biases with a Question-only Branch}

To measure the intensity of language priors in the VQA models, one effective approach is to train a simple model using the question modality alone~\cite{Antol_2015_ICCV,Goyal_2017_CVPR}. 
Due to the large amount of statistical regularities that can be leveraged from the language modality, the question-only model can be a comparable baseline. Technically, the question-only branch is added alongside the base VQA model using MLP layers. 
This question-only branch takes the question embedding $\mathbf{q_i}$ from the question encoder $e_q$. 
In order to avoid the instability brought by the adversarial scheme~\cite{grand2019adversarial}, the question-only model is trained with the objective of minimizing the standard cross entropy loss, which is formalized as: 
\begin{equation}
  \label{con:loss_qo}
  \begin{aligned}
  \mathcal{L}_{QO}
  =-\frac{1}{N} \sum_{i=1}^{N} \log \left(softmax\left(f_{QO}\left(\mathbf{q}_{i}\right)\right)\right) \left[a_{i}\right]
  \end{aligned}
\end{equation}where the $f_{QO}$ denotes the output from this question-only branch as a mapping. Since the question-only branch is trained to predict the correct answer as well as possible, the distribution of $f_{QO}$ can well represent the intensity of language bias. 

\subsection{Reducing Language Biases by Reshaping the VQA Objective Function}

In this section, we specifically introduce how to reduce language biases from the feedback of the question-only branch. First we compute a softmax probability over the outputs from question-only branch. Then we index out the ground-truth answer to generate $\alpha_i$, representing the language bias of the given question:

\begin{equation}
\begin{aligned}
   \alpha_{i}&=softmax\left(f_{QO}\left(\mathbf{q}_{i}\right)\right)\left[a_{i}\right] 
   = \frac{\exp \left(f_{QO}\left(\mathbf{q}_{i}\right)\right)\left[a_{i}\right]}{\sum_{j=1}^{\left | \mathcal{A} \right |} \exp \left(f_{QO}\left(\mathbf{q}_{i}\right)\right)\left[a_{j}\right]}
\end{aligned}
\end{equation}
As $\alpha_i$ increases, the model is more biased toward the language modality. Hence, we adopt this representation $\alpha_i$ to formalize a modulating factor $\beta_i$, which can dynamically adjust the classical VQA objective function (Equation~\ref{con:loss_vqa}) by modifying the loss weights of each training sample. The modulating factor $\beta_i$ is defined as:

\begin{equation}
  \beta_i = \left(1-\alpha_{i}\right)^{\gamma} , \gamma \geq 0
\end{equation}where the $\gamma$ is a tunable hype-parameter for the modulating factor. As $\gamma$ increases, the $\beta_i$ should decrease, so the impact of the modulating factor $\beta_i$ down-weighting the loss function should be likewise increased. Then this modulating factor is combined with the classical VQA objective function (Equation~\ref{con:loss_vqa}) as follows:

\begin{equation}
  \begin{aligned}
    \mathcal{L}_{LPF}
    =-\frac{1}{N} \sum_{i=1}^{N} \underbrace{\left(1-\alpha_{i}\right)^{\gamma}}_{=\beta_i} \log \left(softmax\left(f_{VQA}\left(\mathbf{v}_{\mathbf{i}}, \mathbf{q}_{i}\right)\right)\right) \left[a_{i}\right] 
  \end{aligned}
\end{equation}
There are two properties of this LPF objective function. 
Firstly, when the input sample is not well answered by the question-only branch, the $\alpha_i$ is small so the modulating factor $\beta_i$ is near 1, therefore the loss is hardly affected. 
As $\alpha_i$ grows to 1, the factor $\beta_i$ goes down to 0 and consequently the loss for this most biased sample is significantly down-weighted. 
To better understand the impact of LPF on reducing language biases, we illustrate this with a concrete informal example shown in the Fig.~\ref{fig:banana_example}. 
Assume that the total amount of training samples corresponding to the question type \emph{“What color are the bananas?"} is \emph{n}. Here we set $\gamma$ to 1 for simplicity. 
On the one hand, due to the imbalanced data distribution, the $\alpha_i$ of \emph{“yellow"} bananas in this example is close to 1. 
So the modulating factor $\beta_i$ goes near to 0 and thus the loss for these kinds of overly language bias samples is down-weighted (from \textbf{$28.56 \cdot n \times 10^{-2}$} to \textbf{$5.426 \cdot n \times 10^{-2}$}). 
On the other hand, the $\alpha_i$ of \emph{“green"} banana sample is low, so the modulating factor $\beta_i$ is near 1 and thus the loss for this type of samples is barely affected (from \textbf{$9.16 \cdot n \times 10^{-2}$} to \textbf{$7.969 \cdot n \times 10^{-2}$}). 
The classical biased VQA training objective is re-balanced by the LPF. 
With dynamic feedback from the question-only branch, the LPF enables the VQA model to keep unbiased learning from both modalities to predict answer rather than the language bias. 

Finally, we construct our total loss $L_{total}$ by summing the $L_{QO}$ and $L_{LPF}$ together in the following equation:

\begin{equation}
  \label{con:loss_total}
  \mathcal{L}_{total} = \mathcal{L}_{LPF} + \mathcal{L}_{QO}
\end{equation} 
We jointly optimize the parameters of the base VQA model and the question-only branch in an end-to-end training scheme. 
To prevent the question encoder $e_q$ from straightly learning language biases, gradients calculated from $\mathcal{L}_{QO}$ are not backpropagated to the question encoder~\cite{NIPS2019_8371}. 
By this means, the stronger the question-only branch is, the more the unwanted language priors in the VQA model is captured and removed.

\section{Experiments}
\vspace{-0.5em}
\subsection{Datasets and implementation details}

We validate the effectiveness of our proposed method on the VQA-CP v2 dataset~\cite{Agrawal_2018_CVPR}, the diagnostic benchmark for assessing the VQA model’s generalizability where the answer distributions of training and test splits are significantly different. The dataset contains about 121K images, 438K questions, and 4.4M answers for training, together with about 98K images, 220K questions, and 2.2M answers for testing. 
The results on the validation split of the original VQA v2~\cite{Goyal_2017_CVPR} dataset are also reported for completeness. 
We follow ~\cite{Agrawal_2018_CVPR} to use the standard VQA evaluation metric~\cite{Antol_2015_ICCV} and report accuracy according to different question types. 
For fair comparisons, we adopt the widely-used UpDn~\cite{Anderson_2018_CVPR} model\footnote{\url{https://github.com/hengyuan-hu/bottom-up-attention-vqa}} and do all the same data preprocessing steps. 
Our question-only branch is implemented as a 3-layer MLP of size (2048, 2048, 3000) with ReLU activations. We jointly train the base VQA model and question-only branch end-to-end with parameters initialized from scratch under 
$\mathcal{L}_{total}$
for 21 epochs. 
The Adam optimizer~\cite{kingma2014adam} is used with a learning rate 3e-4 and a batch size of 256. 
After training, the question-only branch is removed to maintain the original VQA model unchanged. 
\vspace{-2em}
\subsection{Results and Analysis}

\subsubsection{Comparison with the SOTAs}

The results of the proposed LPF and other state-of-the-art approaches on the VQA-CP v2 are listed in Table~\ref{table:sota}. We group them into: 
1) baseline VQA models, including GQA~\cite{Agrawal_2018_CVPR} and UpDn~\cite{Anderson_2018_CVPR}. 
2) Extra annotation based methods, including HINT~\cite{Selvaraju_2019_ICCV} and SCR~\cite{NIPS2019_9066}.
3) Data-augmentation based methods, including SSL (CE)~\cite{zhu2020overcoming} and CSS~\cite{chen2020counterfactual}.
4) Non-annotation based methods, including AdvReg~\cite{NIPS2018_7427}, GRL~\cite{grand2019adversarial}, RUBi~\cite{NIPS2019_8371}, VGQE~\cite{gouthamanreducing}, DLR~\cite{jing2020overcoming} and LMH~\cite{clark2019don}. 
Though data-augmentation based methods outperform others by large margins, they change the training priors which makes it hard to evaluate whether VQA models are still driven by memorizing priors. So we only report their results for reference.
The results show that LPF brings significant improvement (\textbf{$+$15.85\%}) over the baseline UpDn model. 
Note that LPF surpasses RUBi by a large margin(\textbf{$+$11.11\%}).
This is because RUBi is still affected by the imbalance training objective, and LPF alleviates this problem by reshaping the global training objective using language priors.
We find that LPF particularly improves VQA performance on \emph{“Yes/No"} questions (\textbf{$+$43.40\%})
because this type of question is susceptible to the influence of language priors.
Our approach also improves the accuracy over the UpDn model (\textbf{$+$3.59\%}) for the question type \emph{“Other"}, while the \textbf{AdvReg}, \textbf{GRL} and \textbf{RUBi} methods experience an obvious performance drop (\textbf{$-$7.50\%}, \textbf{$-$2.22\%} and \textbf{$-$3.37\%}). 
Besides, our method significantly outperforms the \textbf{HINT} and \textbf{SCR} methods, which require extra annotations~\cite{das2017human}. 

\setlength{\tabcolsep}{4pt}
\begin{table}[t]

\begin{center}
\resizebox{\linewidth}{!}{
\begin{tabular}{lllllllllll}
\hline\noalign{\smallskip}
 Model & & \multicolumn{4}{c}{VQA-CP v2 \emph{test}} && \multicolumn{4}{c}{VQA v2 \emph{val}}\\ 
 & $\gamma$ & Overall & Yes/No & Number & Other && Overall & Yes/No & Number & Other\\
\noalign{\smallskip}
\hline
\noalign{\smallskip}

GQA~\cite{Agrawal_2018_CVPR}  && 31.30 & 57.99 & 13.68 & 22.14 && 48.24 & 72.03 & 31.17 & 34.65 \\
UpDn~\cite{Anderson_2018_CVPR}  && 39.49 & 45.21 & 11.96 & 42.98 && 63.48 & 81.18 & 42.14 & 55.66 \\
\hline
UpDn+HINT~\cite{Selvaraju_2019_ICCV} && 46.73 & 67.27 & 10.61 & 45.88 && 63.38 & 81.18 & 42.99 & 55.56 \\
UpDn+SCR~\cite{NIPS2019_9066} && 49.17 & 71.55 & 10.72 & \textbf{47.49} && 62.20 & 78.90 & 41.40 & 54.30 \\
\hline
UpDn+SSL(CE)~\cite{zhu2020overcoming} && 52.63 & 87.75 & 26.40 & 41.42 && 63.73 & \multicolumn{1}{c}{-} & \multicolumn{1}{c}{-} & \multicolumn{1}{c}{-} \\
UpDn+CSS~\cite{chen2020counterfactual} && 58.95 & 84.37 & 49.42 & 48.21 && 59.91 & 73.25 & 39.77 & 55.11 \\
\hline
UpDn+AdvReg~\cite{NIPS2018_7427} && 41.17 & 65.49 & 15.48 & 35.48 && 62.75 & 79.84 & 42.35 & 55.16 \\
UpDn+GRL~\cite{grand2019adversarial} && 42.33 & 59.74 & 14.78 & 40.76 && 51.92 & \multicolumn{1}{c}{-} & \multicolumn{1}{c}{-} & \multicolumn{1}{c}{-} \\
UpDn+RUBi~\cite{NIPS2019_8371} && 44.23 & 67.05 & 17.48 & 39.61 && 61.16 & \multicolumn{1}{c}{-} & \multicolumn{1}{c}{-} & \multicolumn{1}{c}{-} \\
UpDn+VGQE~\cite{gouthamanreducing} && 48.75 & \multicolumn{1}{c}{-} & \multicolumn{1}{c}{-} & \multicolumn{1}{c}{-} && 64.04 & \multicolumn{1}{c}{-} & \multicolumn{1}{c}{-} & \multicolumn{1}{c}{-} \\
UpDn+DLR~\cite{jing2020overcoming} && 48.87 & 70.99 & 18.72 & 45.57 && 57.96 & 76.82 & 39.33 & 48.54 \\
UpDn+LMH~\cite{clark2019don} && 52.01 & 72.58 & \textbf{31.11} & 46.96 && 56.34 & 65.05 & 37.63 & 54.68 \\
\hline
\textbf{UpDn$+$LPF(ours)} & 1 & 51.57 & 87.33 & 12.25 & 43.61 && 62.63 & 79.51 & 42.90 & 55.02 \\
\textbf{UpDn$+$LPF(ours)} & 5 & \textbf{55.34} & \textbf{88.61} & \underline{23.78} & 46.57 && 55.01 & 64.87 & 37.45 & 52.08 \\
\hline
\end{tabular}
}
\vspace{-1.5em}
\caption{Comparison of the performance on VQA-CP v2 \emph{test} and VQA v2 \emph{val} set with the state-of-the-art systems. 
\vspace{-3em}}
\label{table:sota}
\end{center}
\end{table}
\setlength{\tabcolsep}{1.4pt}

\subsubsection{Results on biased VQA dataset}

In Table~\ref{table:sota}, we also report the results on the VQA v2 dataset containing strong language prior. Different from the VQA-CP, the \emph{train} and \emph{val} sets of VQA v2 dataset follow the same distribution, so the performance of most previous de-biasing methods are more or less dropped. 
The hype-parameter $\gamma$ of LPF makes a trade-off between reducing language biases and answering questions. 
So the LPF approach maintains competitive performance on the VQA v2 when the $\gamma=1$, and performs best on the VQA-CP v2 when $\gamma=5$. 

\subsubsection{LPF is model-agnostic} 

To demonstrate that our method works well on different VQA models, we also build LPF framework on the 
BAN~\cite{NIPS2018_7429} 
and the 
S-MRL~\cite{NIPS2019_8371}. 
From the results depicted in Table~\ref{table:agnostic}, the significant improvements in all different VQA models(\textbf{$\sim$+13\%} to \textbf{$\sim$+16\%}) demonstrate that LPF is model-agnostic.

\setlength{\tabcolsep}{4pt}

\begin{table}[t]
\begin{center}

\resizebox{77mm}{!}{
\begin{tabular}{@{}lccccc@{}}
\hline
\multicolumn{1}{c}{Model} & Overall        & Y/N            & Number         & Other          & Gap$\Delta\uparrow$                  \\ 
\hline
S-MRL~\cite{NIPS2019_8371}                     & 38.46          & 42.85          & 12.81          & \textbf{43.20} & \multicolumn{1}{l}{} \\
S-MRL+RUBi~\cite{NIPS2019_8371}                & 47.11          & 68.65          & 20.28          & 43.18          & $+$8.65                 \\
\textbf{S-MRL+LPF(ours)}  & \textbf{53.38} & \textbf{88.06} & \textbf{25.00} & 42.99          & \textbf{$+$14.92}       \\ 
\hline
UpDn~\cite{Anderson_2018_CVPR}                      & 39.49          & 45.21          & 11.96          & 42.98          & \multicolumn{1}{l}{} \\
UpDn+RUBi~\cite{NIPS2019_8371}                 & 44.23          & 67.05          & 17.48          & 39.61          & $+$4.74                 \\
\textbf{UpDn+LPF(ours)}   & \textbf{55.34} & \textbf{88.61} & \textbf{23.78} & \textbf{46.57} & \textbf{$+$15.85}       \\ 
\hline
BAN~\cite{NIPS2018_7429}                      & 37.03          & 41.55          & 12.43          & \textbf{41.40}          & \multicolumn{1}{l}{} \\

\textbf{BAN+LPF(ours)}   & \textbf{50.76} & \textbf{88.13} & \textbf{18.59} & 40.03 & \textbf{$+$13.73}       \\ 
\hline
\end{tabular}
}
\caption{Performance (\%) on the VQA-CP v2 test set based on different VQA models. Gap$\Delta\uparrow$ represents the accuracy improvements of the \emph{“Overall"} question type.\vspace{-3em}}
\label{table:agnostic}
\end{center}
\end{table}
\setlength{\tabcolsep}{1.4pt}

\subsubsection{Diffenrent Varients of LPF}
\label{sec:var_exp}
The original $\alpha_i$ in LPF is a softmax probability $P_{f_{QO}}(a_i|Q_i)$ dynamically learnt from the Question-only model. To further discuss the effectiveness of our proposed LPF, we explore different ways to calculate $\alpha_i$ in LPF objective function: 
1) \textbf{UpDn+Precomputing}: 
In order to explore how important it is to learn a probability that represents the language bias,
we conduct the experiments of precomputing the empirical answer distribution per question type $P(a_i|Q_{type})$ from the training set as $\alpha_i$, \emph{i.e.}, $\alpha_i=P(a_i|Q_{type})$. 
2) \textbf{UpDn+Focal}: 
To explore whether the LPF is superior than the widely used reweighting mechanism, such as focal loss~\cite{Lin_2017_ICCV},
we extend the original focal loss to the multi-class case and apply it to VQA. 
In practice, we compute a softmax probability over the outputs from VQA model itself and index out the ground-truth answer, \emph{i.e.}, $\alpha_{i}=softmax\left(f_{VQA}\left(\mathbf{v}_{i},\mathbf{q}_{i}\right)\right)\left[a_{i}\right]$. 
In both experiments, the $\beta_i$ is set by the changed $\alpha_{i}$ and the $\gamma$ is set to 1, \emph{i.e.}, $\beta_{i}=1-\alpha_i$. 
As results shown in Table~\ref{table:varients}, \textbf{UpDn+Precomputing} brings some performance improvements ($+$0.55) from the UpDn models. Nevertheless, dynamically learning the language bias shows much favourable performance (\textbf{$+$12.08}) rather than precomputing. 
On the other hand, \textbf{UpDn+Focal} experiences some performance drops ($-$0.97). We analyze that the output of VQA model itself comprises visual information as confounder so it can't well represent the language bias. 
Taking a simple auxiliary question-only branch is more appropriate for estimation of language bias. 

\setlength{\tabcolsep}{4pt}
\begin{table}[t]
\begin{center}
\resizebox{77mm}{!}{
\begin{tabular}{@{}llllll@{}}
\hline
Model                                                  & Overall & Yes/No & Number & Other & Gap$\Delta\uparrow$ \\ 
\hline
UpDn~\cite{Anderson_2018_CVPR} & 39.49   & 45.21  & 11.96  & 42.98 &                     \\ \midrule
UpDn+Precomputing                                      & 40.04   & 44.81  & 11.73  & \textbf{45.31} & $+$0.55               \\
UpDn+Focal                                             & 38.52   & 42.38  & \textbf{12.38}  & 43.67 & $-$0.97               \\
\textbf{UpDn+LPF}                                               & \textbf{51.57}   & \textbf{87.33}  & 12.25  & 43.61 & \textbf{$+$12.08}              \\ 
\hline
\end{tabular}
}
\caption{Results of different varients of LPF on the VQA-CP v2 \emph{test} split. 
\vspace{-3em}}
\label{table:varients}
\end{center}
\end{table}
\setlength{\tabcolsep}{1.4pt}

\subsubsection{LPF Results in Less Biased Answer Distributions} 
\begin{figure}[t]
 \centering
 \subfigure{
    \includegraphics[width=0.5\columnwidth]{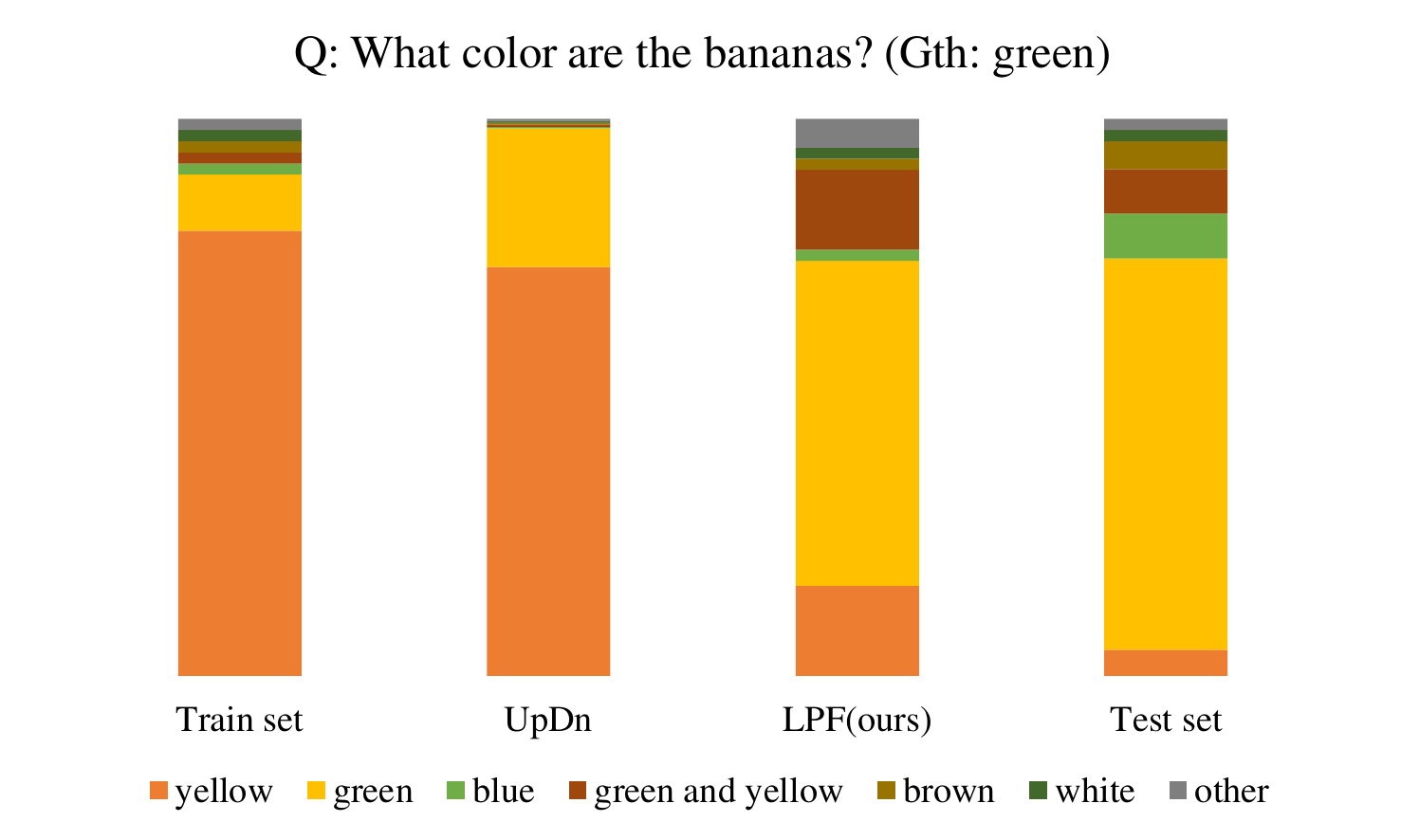} 
    \includegraphics[width=0.5\columnwidth]{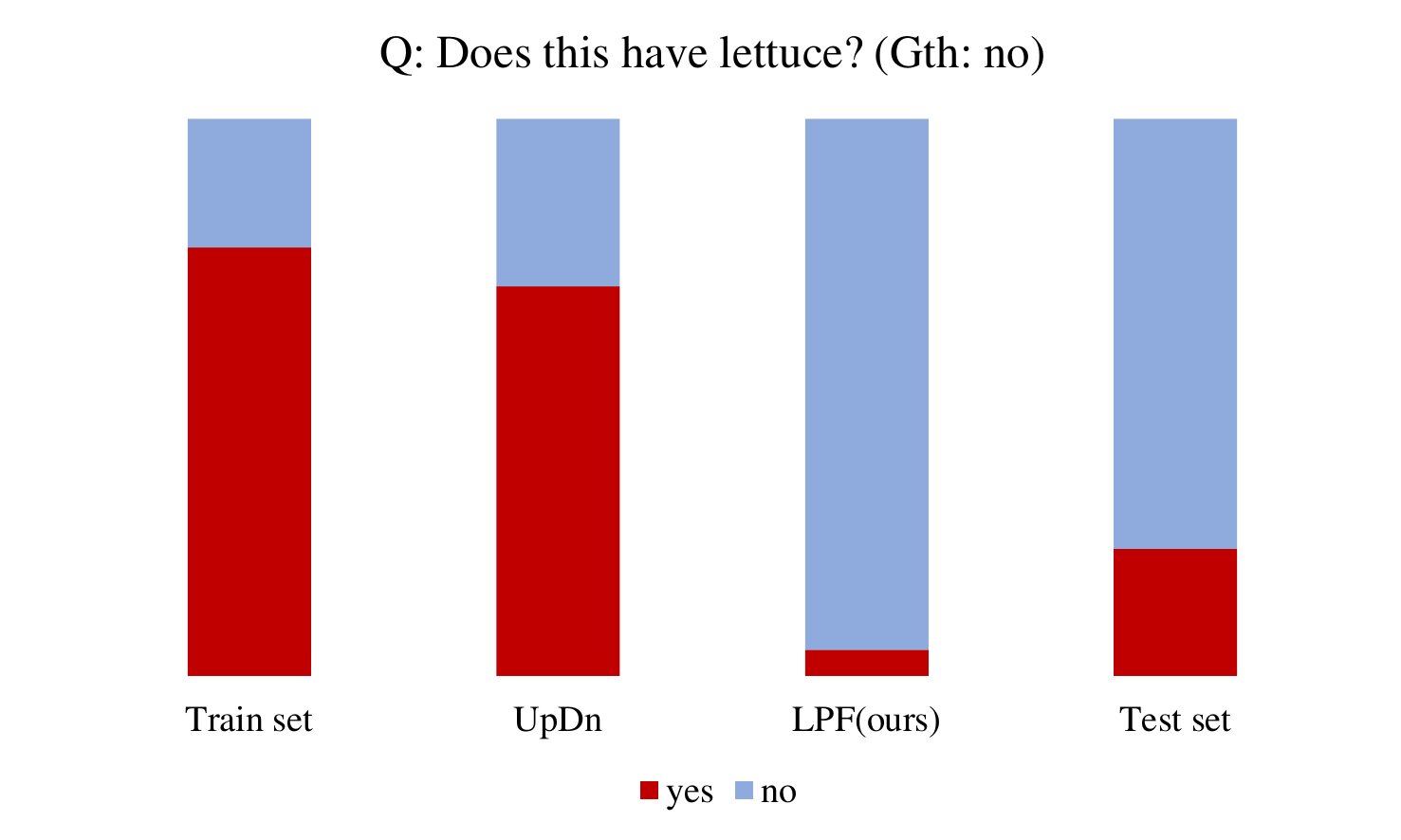} 
  }
  \vspace{-1em}
  \caption{The answer distributions on the VQA-CP v2.\vspace{-1em}}
  \label{fig:distribution}
\end{figure}

Fig.~\ref{fig:distribution} shows answer distributions for VQA-CP v2 \textit{train} and \textit{test} sets, the UpDn and the LPF of two question examples: “\emph{What color are the bananas?}” and “\emph{Does this have lettuce?}”. It shows that the original UpDn model tends to overfit the answer distribution of training set and predicts the frequently-used answer in it rather than reasoning from the visual content. 
In contrast, the LPF can well recover the answer distribution in the test set, which demonstrates that LPF enables the model to be more grounded on the image content. 
\vspace{-1em}

\begin{figure}[t]
  \centering
  \includegraphics[width=\linewidth]{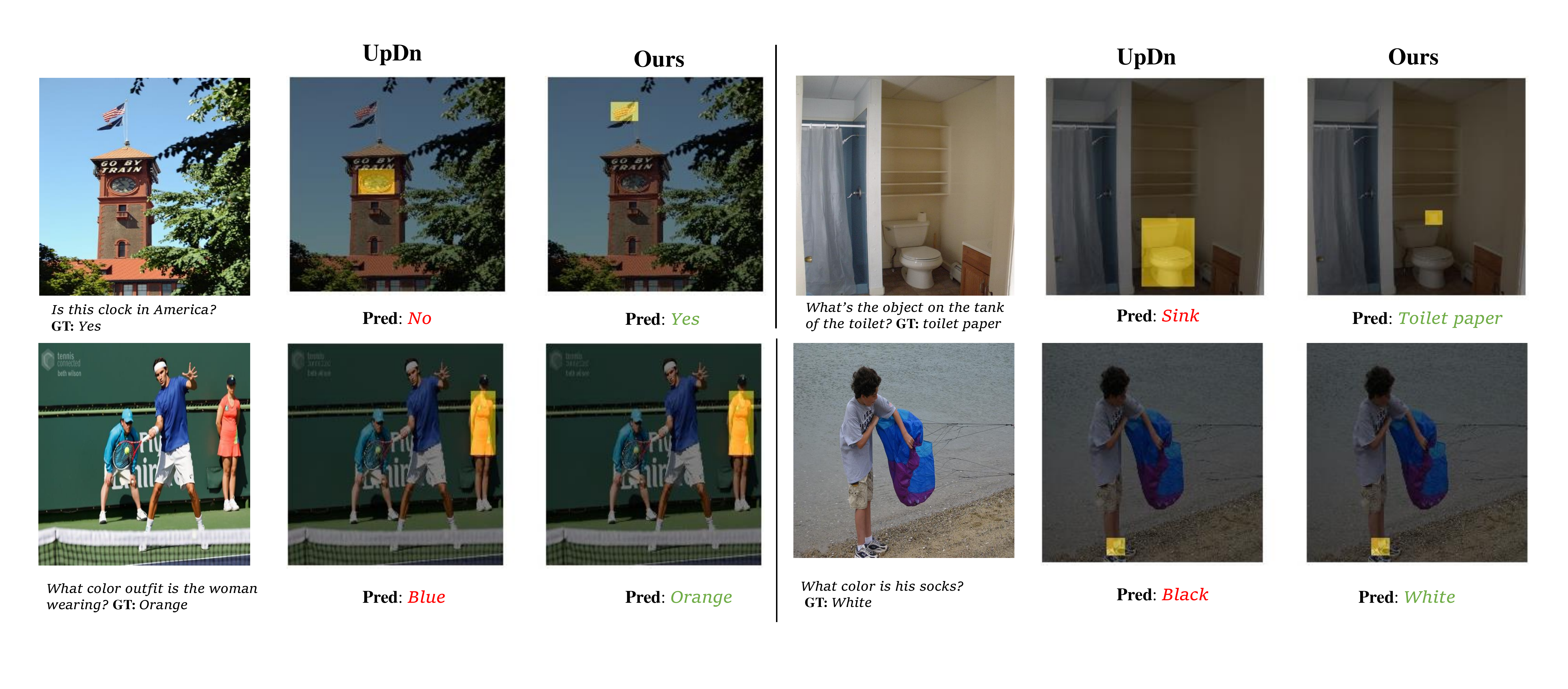}
  \vspace{-2em}
  \caption{Qualitative comparison results of the original UpDn model and ours on VQA-CP v2 \emph{test} split. 
  \vspace{-1em}
  }
  \label{fig:visualize}
\end{figure}

\subsection{Qualitative Analysis}

We display some qualitative comparisons between UpDn and our method in Fig.~\ref{fig:visualize}. 
As we can see in the upper row, LPF helps model to attend to a more reasonable visual region (\emph{“American flag"} and \emph{“toilet paper"}) to deduce the right answer. 
The results reveal that our method can improve the “visual" ability of the base model. 
In the bottom row, the baseline UpDn makes mistakes even it attends to the right visual region, which may be caused by the language biases. Instead, LPF enables the model to be more robust to “reason" about the relevant image regions before answering the question.

\vspace{-1em}
\section{Conclusions}

In this work, we proposed a generic training method named LPF to tackle the language bias problem in VQA. The LPF assigns a dynamic weight to each training sample and re-balances the total VQA loss based on the feedback from the question-only branch.
The experimental results demonstrate that our method outperforms other strong baselines.
In future work, we will explore whether our approach deals with uni-modal biases on other multi-modal tasks.

\bibliographystyle{ACM-Reference-Format}
\balance
\bibliography{sample-base}




\end{document}